\documentclass[letterpaper]{article} 
\PassOptionsToPackage{table}{xcolor}
\usepackage[preprint]{aaai2027} 
\usepackage[hyphens]{url} 
\usepackage{graphicx} 
\usepackage{natbib} 
\usepackage{caption} 
\usepackage{subcaption}
\usepackage{amsmath,amsfonts}
\usepackage{adjustbox}
\usepackage{booktabs}
\usepackage{multirow}
\usepackage{makecell}

\usepackage{algorithm}
\usepackage{algorithmic}

\title{Route-and-Generate: KL-Regularized Router Improvement for Reasoning Routing in LLMs}

\author{
    Tien-Phat Nguyen\textsuperscript{\rm 1}\equalcontrib\corresponding,
    Truong Nguyen\textsuperscript{\rm 1}\equalcontrib,
    Thanh-Thien Le\textsuperscript{\rm 2},
    Hoang Phan\textsuperscript{\rm 3},\\
    Thien Huu Nguyen\textsuperscript{\rm 4},
    Linh Ngo Van\textsuperscript{\rm 1},
    Trung Le\textsuperscript{\rm 2}
}
\affiliations{
    \textsuperscript{\rm 1}Hanoi University of Science and Technology, Vietnam \qquad
    \textsuperscript{\rm 2}Monash University, Australia\\
    \textsuperscript{\rm 3}New York University, USA \qquad
    \textsuperscript{\rm 4}University of Oregon, USA
}

\begin{document}
\maketitle
\begin{abstract}
Large language models often apply complex reasoning indiscriminately, leading to overthinking on simple tasks and underthinking on harder ones. We present \textbf{Route-and-Generate (Rouge)}, a hierarchical training framework that enables LLMs to adaptively decide when and how much reasoning is required for each input. Our approach decouples reasoning effort allocation from answer generation by introducing a routing policy that allocates a fixed sampling budget across multiple reasoning formats, while the base model is optimized with GRPO using per-sample rewards. To promote efficient reasoning, we employ a rank-based reward that favors shorter correct solutions, and update the routing policy via a closed-form KL-regularized policy improvement step distilled into a neural router. This design encourages concise answers on easy problems while reserving longer reasoning for harder ones, achieving improved accuracy–efficiency trade-offs across multiple reasoning benchmarks.
\end{abstract}

\section{Introduction}

Large reasoning models \cite{jaech2024openai, r1} can solve problems that would otherwise be out of reach by generating explicit intermediate reasoning. Unfortunately, this extra reasoning is expensive: each additional token adds latency and cost, and the gap between ``enough'' and ``too much'' reasoning widens as models scale. In practice, we often see a frustrating pattern---models strain on routine inputs with needless multi-step traces, while still under-allocating depth on the hardest questions where it matters most.

This failure mode persists because most current training paradigms \cite{wu2024how, aggarwal2025l1} optimize primarily for accuracy, so a single policy must simultaneously learn \emph{how} to produce the correct answer and \emph{what} reasoning strategy or format to use. While reinforcement learning can raise pass rates, it also creates a tension between success and efficiency: longer trajectories often improve accuracy \cite{chen2024not}, but only up to a point, after which extra tokens buy diminishing returns. As a result, policies drift toward verbosity even when the marginal gains are small.

We argue that the goal of separating answer generation from the choice of an input-specific reasoning strategy is better captured by a sharper question: \emph{given a fixed compute budget, how should a model allocate reasoning across alternative strategies to maximize correctness?} This framing compels a disciplined choice about \emph{when} additional reasoning is worth its cost, rather than purely optimizing for correctness, and it cleanly separates two roles—selecting a strategy and executing it to produce an answer—so that ``reasoning effort'' becomes a controllable variable rather than an incidental byproduct of training.


Based on this perspective, we propose \textbf{Route-and-Generate} (\textbf{Rouge}), a training framework that decouples reasoning strategy selection from answer generation. Instead of committing to a single reasoning trajectory, \textbf{Rouge} distributes computation across multiple reasoning formats and generates candidate solutions accordingly. This separation alleviates the core tension between accuracy and efficiency because the model no longer needs to encode both how much reasoning to use and how to produce the answer within a single generation policy. Instead, reasoning effort is decided explicitly at the routing stage, allowing the model to reserve expensive reasoning for inputs that truly require it while falling back to concise strategies when additional computation would be wasteful. Strategy selection, however, remains challenging: prior work has explored choosing among reasoning modes \cite{wu2025armadaptivereasoningmodel}, including Mixture-of-Experts selectors \cite{si-etal-2023-getting}, but such routers typically rely on explicit supervision or predefined preferences over which format to use. In contrast, we formulate routing supervision as an optimization problem and derive a closed-form solution for assigning soft routing targets.

We summarize our contributions as follows:
\begin{itemize}
    \item \textbf{Route-and-Generate (Rouge).} We introduce a training framework that decouples reasoning strategy selection from reasoning generation, enabling explicit control over both reasoning format and computational effort.

    \item \textbf{Theoretically grounded KL-Regularized Routing.} We formulate reasoning strategy supervision as a KL-regularized optimization problem and derive a closed-form solution for improving routing without manual labels.

    \item \textbf{Extensive experiments.} We validate Rouge across multiple reasoning benchmarks and analysis, demonstrating improved efficiency without sacrificing accuracy.
\end{itemize}

\section{Related Work}

\paragraph{Reasoning in Large Language Models.}
Recent large language models have demonstrated strong reasoning capabilities through explicit intermediate reasoning mechanisms, including chain-of-thought prompting, programmatic reasoning, and iterative self-reflection \citep{wei2022chain, li2025codeio, r1}. These approaches enable models to decompose complex problems into intermediate steps and have shown substantial improvements on mathematical, logical, and symbolic reasoning benchmarks. However, explicit reasoning typically incurs increased generation length and computational cost, and is often applied uniformly across inputs regardless of task difficulty, leading to inefficiencies on simpler problems \citep{chen2024not, sui2025stop}.

\paragraph{Reinforcement Learning for Reasoning.}
Reinforcement learning with verifiable or rule-based rewards has emerged as an effective paradigm for improving reasoning capabilities in language models \citep{cobbe2021training, shao2024deepseekmath, r1}. By directly optimizing task-level correctness, these methods encourage the emergence of complex reasoning behaviors, such as long-horizon planning and self-verification. However, when reasoning structure, length, and generation are optimized jointly, policy optimization methods tend to favor more expressive and verbose reasoning processes, as longer reasoning trajectories often achieve higher success probability under accuracy-driven objectives. This coupling introduces a bias toward excessive reasoning and motivates recent work on improving reasoning efficiency without sacrificing performance \citep{cheng2024compressed, fan2025missing}.

\paragraph{Adaptive and Efficient Reasoning.}
To mitigate unnecessary computation, several approaches explore adaptive and efficient reasoning strategies, including multiple reasoning formats, length-controlled generation, and efficiency-aware reinforcement learning objectives \citep{wu2025armadaptivereasoningmodel, hou2025thinkprune, aggarwal2025l1}. These methods reduce average computational cost by encouraging concise reasoning or selecting among predefined reasoning modes. However, they typically regulate reasoning effort through discrete format selection or global length penalties applied to the generated sequence, and do not explicitly model how computational resources should be allocated across different reasoning strategies within a single decision process.

\paragraph{Length and Budget Aware Reinforcement Learning.}
More broadly, controlling computational cost in reinforcement learning has been studied through cost and length-aware objectives that explicitly penalize excessive computation during training or inference. In the context of language models, several works incorporate output length, token budget, or computation cost directly into reinforcement learning objectives, encouraging models to trade off task performance against reasoning length or generation cost \citep{wu2024how, o1prunner,aggarwal2025l1, hou2025thinkprune}. While effective at reducing unnecessary computation, these approaches typically impose global penalties or constraints on the generated sequence, implicitly coupling the decision of reasoning effort with reasoning execution. In contrast, our work is inspired by this line of research but focuses on reasoning in language models by treating different reasoning strategies as alternative consumers of a shared computation budget, and by explicitly separating the decision of reasoning effort from the generation of reasoning content.

\section{Methodology}
\paragraph{Method Overview.}
Rouge separates reasoning-effort allocation from answer generation using two components: a format router \(g_{\theta}(f \mid q)\), which predicts a distribution over reasoning formats for each question, and a format-conditioned generator \(\pi_{\phi}(y \mid q,f)\), which produces candidates under the selected format templates. The router allocates a fixed generation budget across formats, the generator samples format-specific candidates, and these candidates are scored by an efficiency-aware reward that favors concise correct solutions. The remainder of this section first discusses the motivation for this decomposition, then presents routing and candidate generation, followed by the reward design, the router objective, and the full training algorithm.
\subsection{Motivation}
\label{sec:motivation}

We consider a supervised reasoning setting where each training example consists of a question paired with reference signals, such as a ground-truth answer, unit tests, or verifier inputs. Following prior work, we assume that a small set of reasoning formats is sufficient to cover most problem instances,
\[
\mathcal{F}=\{\text{\texttt{Direct}},\ \text{\texttt{Short-CoT}},\ \text{\texttt{Code}},\ \text{\texttt{Long-CoT}}\},
\]
where each format \(f \in \mathcal{F}\) is associated with a prompt template \(T_f(\cdot)\).

ARM~\cite{wu2025armadaptivereasoningmodel} trains a single model to use multiple reasoning formats and mitigates format collapse by encouraging exploration of under-sampled formats during training. However, this learning signal has two limitations. First, ARM treats formats as discrete categories and does not control cost variation within a format. In practice, outputs under the same format, especially \texttt{Code} or \texttt{Long-CoT}, can differ substantially in length and computational depth, so format-level diversity alone does not guarantee efficiency. Second, ARM does not explicitly learn a principled task-conditioned format policy. To avoid format collapse, it relies on a diversity-promoting reward adjustment whose effect decays over training. As a result, any relationship between question difficulty and format choice is induced only indirectly by the training dynamics, rather than being optimized as an explicit accuracy--efficiency trade-off.

These limitations define the scope of our problem. We study \emph{adaptive
reasoning-format selection}: the decision variable is a semantic reasoning mode,
not a target token count. Output length is an outcome of executing the selected
format and enters our reward only to distinguish efficient from wasteful
realizations. Under a fixed backbone, format set, and sampling budget, the system
must therefore make two separate decisions: \textit{which} format suits the
input, and \textit{how} to execute it efficiently. \textbf{Rouge} explicitly
separates these decisions through its router and generator.
\subsection{Format Routing and Candidate Generation}
\label{sec:allocation_generation}

Given a question $q$, a neural \emph{format router} predicts a distribution over 
reasoning formats,
\[
g_{\theta}(f \mid q), \qquad f \in \mathcal{F},
\]
which determines how the available computation budget should be allocated across 
alternative reasoning strategies. Concretely, the router is implemented as a 
lightweight encoder-based classifier that maps the input question to a categorical 
distribution over formats.

Under a fixed sampling budget of $K$ generations per question, we convert this 
distribution into a discrete allocation by sampling
\[
\mathbf{n}(q)=\{n_f\}_{f\in\mathcal{F}} \sim 
\mathrm{Multinomial}\!\big(K,\, g_{\theta}(\cdot \mid q)\big),
\]
so that $\sum_{f\in\mathcal{F}} n_f = K$. This produces a question-dependent 
mixture of reasoning strategies, allowing simpler inputs to receive more concise 
formats while reserving heavier reasoning for harder problems.

Conditioned on the resulting allocation, the generator produces $n_f$ candidate 
solutions for each format $f$ using a format-specific prompt template $T_f(q)$. The format signal is communicated
implicitly through the template, i.e.,
\[
\pi_{\phi}(\cdot \mid T_f(q)) \equiv 
\pi_{\phi}(\cdot \mid q, f),
\]
and candidates are sampled as
\[
y_{f,i} \sim \pi_{\phi}\big(\cdot \mid T_f(q)\big), 
\qquad i=1,\dots,n_f.
\]

Aggregating candidates across all formats yields the rollout set
\[
\mathcal{Y}(q)=\bigcup_{f\in\mathcal{F}}\{y_{f,1},\dots,y_{f,n_f}\}.
\]
This design cleanly separates \emph{reasoning-effort allocation} from 
\emph{within-format generation}: the router decides \emph{where} computation is 
spent, while the generator determines \emph{how} solutions are realized under 
each selected reasoning format.

\subsection{Reward Design}
\label{sec:reward_design}

\begin{algorithm}[t!]
\caption{Training Algorithm for Rouge Framework}
\label{alg:train}
\begin{algorithmic}[1]
\REQUIRE Batch of $(q,R)\sim\mathcal{D}_{\text{train}}$, format list $\mathcal{F}$, budget $K$, reference router $g_{\text{ref}}$, coefficients $(\rho,\beta,\lambda_{\text{LLM}})$
\FOR{each training instance $(q,R)$}
  \STATE Router forward: compute $g_{\theta}(f\mid q)$ for all $f\in\mathcal{F}$
  \STATE Allocate: $\{n_f\}\leftarrow \textsc{Alloc}(K,g_{\theta}(\cdot\mid q))$
  \FOR{each format $f\in\mathcal{F}$}
    \FOR{$i=1$ to $n_f$}
      \STATE Generate $y_{f,i}\sim \pi_{\phi}(\cdot\mid T_f(q))$
      \STATE Compute $r_{\text{LLM}}(q,y_{f,i};R)$ via Eq.~\eqref{eq:r_llm}
    \ENDFOR
    \STATE Aggregate $\hat r_{\text{fmt}}(q,f)$ via Eq.~\eqref{eq:r_fmt}
  \ENDFOR
  \STATE \textbf{Generator update:} take one GRPO step on $\phi$ (Eq.~\eqref{eq:llm_grpo})
  \STATE Compute router target $g^{*}(f\mid q)$ via Eq.~\eqref{eq:router_closed_form}
  \STATE \textbf{Router update:} take one gradient step on $\theta$ minimizing Eq.~\eqref{eq:router_distill}
\ENDFOR
\end{algorithmic}
\end{algorithm}

To jointly guide reasoning correctness and computational efficiency, we use a 
rank-based reward defined over the set of candidates generated for each question.

\paragraph{Relative efficiency reward.}
Let $\text{Acc}(y;R)\in\{0,1\}$ denote correctness under a task-specific verifier 
or reference signal $R$. For a given question $q$, define the set of correct 
candidates
\[
\mathcal{Y}^+(q)=\{y\in\mathcal{Y}(q)\mid \text{Acc}(y;R)=1\}.
\]
We sort $\mathcal{Y}^+(q)$ in ascending order of generation length and denote 
the candidate with rank $k$ (starting from $k=0$) by $y_{(k)}$, so that 
$\text{Len}(y_{(k)}) \le \text{Len}(y_{(k+1)})$ for all valid indices $k$. 
The reward for each candidate is then defined as
\begin{equation}
\label{eq:r_llm}
r_{\text{LLM}}(q,y;R)=
\begin{cases}
\rho + \dfrac{1-\rho}{2^{k}}, 
& \begin{aligned}
\text{if } y = y_{(k)} \\
\text{for some rank } k,
\end{aligned} \\
0, & \text{if } \text{Acc}(y;R)=0.
\end{cases}
\end{equation}
where $\rho\in(0,1)$ ensures a strict margin between correct and incorrect 
responses. This reward assigns the highest value to the shortest correct 
solution and decays exponentially with rank, providing a batch-relative 
efficiency signal that is invariant to absolute length scales.

\paragraph{Format-level reward.}
To update the router’s budget allocation policy, we aggregate candidate-level 
rewards within each reasoning format. Specifically, for format $f$ with allocation 
$n_f$, we define
\begin{equation}
\hat r_{\text{fmt}}(q,f) = \frac{1}{n_f}\sum_{i=1}^{n_f} 
r_{\text{LLM}}(q,y_{f,i};R),
\label{eq:r_fmt}
\end{equation}
i.e., the mean reward across candidates generated under format $f$. This 
aggregation provides a stable estimate of format utility, enabling the router to 
shift computation toward reasoning strategies that yield more efficient correct
solutions.

\subsection{Generator Optimization}
\label{sec:generator_opt}

We optimize the generator policy $\pi_{\phi}$ using GRPO-style policy gradient 
updates~\cite{shao2024deepseekmath}. For each question $q$, advantages are computed by normalizing 
candidate rewards within the rollout set:
\begin{equation}
A(q,f,y)=
\frac{
r_{\text{LLM}}(q,y;R)-\mu_q
}{
\sigma_q+\epsilon
},
\end{equation}
where $\mu_q$ and $\sigma_q$ denote the mean and standard deviation of rewards 
over $\mathcal{Y}(q)$.

The generator is updated by maximizing
\begin{equation}
\label{eq:llm_grpo}
\begin{split}
\mathcal{L}_{\text{LLM}}(\phi)=
-\mathbb{E}\Big[A(q,f,y)\,\log \pi_{\phi}(y\mid q,f)\Big] \\
+\lambda_{\text{LLM}}\,\mathrm{KL}\!\left(\pi_{\phi}\ \|\ \pi_{\phi,\text{ref}}\right),
\end{split}
\end{equation}
where $\pi_{\phi,\text{ref}}$ is a frozen reference policy used for KL 
regularization.
\subsection{Closed-form KL-Regularized Router}
\label{sec:router}

The router acts as a meta-controller that allocates a fixed computation budget 
across reasoning formats by predicting a distribution $g_{\theta}(f\mid q)$. 
Given format-level utilities derived from the relative efficiency reward 
defined above, we update the routing policy via a 
KL-regularized policy improvement step:
\begin{equation}
\label{eq:router_obj}
\begin{aligned}
\max_{g(\cdot\mid q)}\;&
\mathbb{E}_{f\sim g(\cdot\mid q)}
\big[\hat r_{\mathrm{fmt}}(q,f)\big] \\
&-
\beta\,\mathrm{KL}\!\big(
g(\cdot\mid q)\ \|\ g_{\text{ref}}(\cdot\mid q)
\big),
\end{aligned}
\end{equation}
where $\hat r_{\mathrm{fmt}}(q,f)$ denotes the aggregated utility of format $f$ 
for question $q$, and $g_{\text{ref}}$ is a reference routing policy (e.g., uniform 
or warm-started). The regularization coefficient $\beta>0$ controls the trade-off 
between exploiting high-utility formats and maintaining exploration.

This objective admits a closed-form solution:
\begin{equation}
\label{eq:router_closed_form}
g^{*}(f\mid q)\ \propto\ 
g_{\text{ref}}(f\mid q)
\exp\!\left(\frac{\hat r_{\mathrm{fmt}}(q,f)}{\beta}\right),
\end{equation}
which reweights the reference policy according to exponentiated format utilities. 
Intuitively, formats that yield more efficient correct solutions receive higher 
probability mass, while the KL term prevents premature routing collapse.
\textcolor{black}{In practice, we anneal $\beta$ from large to small over
training, encouraging broad exploration early and sharper routing later.}

We distill this improved policy into the parametric router by minimizing
\begin{equation}
\label{eq:router_distill}
\mathcal{L}_{\text{format}}(\theta)=
\mathbb{E}_{q\sim \mathcal{D}_{\text{train}}}
\Big[
\mathrm{KL}\big(g^{*}(\cdot\mid q)\ \|\ g_{\theta}(\cdot\mid q)\big)
\Big].
\end{equation}



\newcommand{\best}[1]{\textbf{#1}}

\begin{table*}[t]
\centering
\renewcommand{\arraystretch}{1.12}

\begin{adjustbox}{max width=\linewidth}
\begin{tabular}{ll*{6}{r r}}
\toprule
\multirow{2}{*}{\textbf{Metric}} &
\multirow{2}{*}{\textbf{Method}} &
\multicolumn{2}{c}{\textbf{AIME 2025}} &
\multicolumn{2}{c}{\textbf{AIME 2024}} &
\multicolumn{2}{c}{\textbf{AMC 2023}} &
\multicolumn{2}{c}{\textbf{MATH}} &
\multicolumn{2}{c}{\textbf{GSM8K}} &
\multicolumn{2}{c}{\textbf{SVAMP}} \\
\cmidrule(lr){3-4}\cmidrule(lr){5-6}\cmidrule(lr){7-8}\cmidrule(lr){9-10}\cmidrule(lr){11-12}\cmidrule(lr){13-14}
& & \textbf{Score} & \textbf{\#Tok}
  & \textbf{Score} & \textbf{\#Tok}
  & \textbf{Score} & \textbf{\#Tok}
  & \textbf{Score} & \textbf{\#Tok}
  & \textbf{Score} & \textbf{\#Tok}
  & \textbf{Score} & \textbf{\#Tok} \\
\midrule

\rowcolor{gray!15}\multicolumn{14}{c}{\best{Qwen2.5 3B}}\\

\midrule

\multirow{5}{*}{\best{Pass@1}}    
                              & SFT & 0.00 & 694 & 0.00 & 35 & 7.50 & 75 & 20.90 & 229 & 35.70 & 145 & 62.30 & 126 \\
                              & O1-Pruner & \best{3.33} & 125 & \best{6.67} & 1401 & 20.00 & 876 & 40.08 & 314 & 81.50 & 253 & 86.33 & 137 \\
                              \cmidrule{2-14}
                              & GRPO & \best{3.33} & 2977 & \best{6.67} & \underbar{2978} & 30.00 & \underbar{2139} & \best{64.36} & 1442 & \best{87.57} & \underbar{525} & \best{92.00} & 413 \\
                              & ARM  & 0.00 & \underbar{2916} & \best{6.67} & 3298 & \underbar{40.00} & 2163 & 63.66 & \underbar{1421} & 87.04 & 535 & \underbar{91.00} & \underbar{365} \\
                              & \textbf{Rouge} & \best{3.33} & \best{2531} & \best{6.67} & \best{2849} & \best{42.50} & \best{1413} & \underbar{64.04} & \best{1202} & \underbar{87.19} & \best{422} & 88.67 & \best{242} \\

\midrule

\multirow{5}{*}{\best{Maj@8}}  
                              & SFT & 0.00 & 537 & 0.00 & 450 & 17.50 & 320 & 27.60 & 231 & 56.00 & 132 & 74.70 & 108 \\
                              & O1-Pruner & \underbar{3.33} & 385 & 6.67 & 928 & 30.00 & 685 & 47.66 & 393 & 86.43 & 250 & 89.67 & 174 \\
                              \cmidrule{2-14}
                              & GRPO & \best{6.67} & 2671 & \underbar{10.00} & 2921 & \best{45.00} & 1909 & \best{72.88} & 1329 & \best{91.05} & 485 & \best{93.33} & 390 \\
                              & ARM  & \underbar{3.33} & \underbar{2298} & \underbar{10.00} & \underbar{2192} & \best{45.00} & \underbar{1428} & \underbar{71.90} & \underbar{1040} & \underbar{89.50} & \underbar{356} & 92.00 & \underbar{243} \\
                              & \textbf{Rouge} & \underbar{3.33} & \best{2042} & \best{13.33} & \best{1992} & \best{45.00} & \best{1252} & 71.38 & \best{884} & \underbar{89.50} & \best{304} & \underbar{92.33} & \best{206} \\

\midrule
\rowcolor{gray!15}\multicolumn{14}{c}{\best{Qwen2.5 7B}}\\
\midrule

\multirow{5}{*}{\best{Pass@1}}
                              & SFT & 0.00 & 1239 & 0.00 & 40 & 12.50 & 91 & 30.40 & 348 & 54.40 & 184 & 76.00 & 126 \\
                              & O1-Pruner & 3.33 & 911 & 3.33 & 921 & 35.00 & 963 & 52.78 & 693 & 79.38 & 306 & 83.67 & 207 \\
                              \cmidrule{2-14}
                              & GRPO & \underbar{6.67} & \underbar{3156} & \underbar{16.67} & 3159 & \underbar{55.00} & \underbar{2112} & \underbar{77.90} & \underbar{1232} & \underbar{92.27} & \underbar{445} & \underbar{94.00} & 401 \\
                              & ARM & \best{16.67} & \underbar{3156} & 13.33 & \underbar{3083} & \underbar{55.00} & 2126 & \best{77.96} & 1240 & \underbar{92.27} & 448 & \underbar{94.00} & \underbar{330} \\
                              & \textbf{Rouge} & \underbar{6.67} & \best{2778} & \best{20.00} & \best{3024} & \best{57.50} & \best{2031} & 77.36 & \best{1106} & \best{92.42} & \best{347} & \best{94.67} & \best{260} \\

\midrule

\multirow{5}{*}{\best{Maj@8}}
                              & SFT & 0.00 & 1023 & 0.00 & 522 & 27.50 & 381 & 42.40 & 361 & 79.40 & 185 & 88.00 & 141 \\
                              & O1-Pruner & 3.33 & 1058 & 10.00 & 937 & 52.50 & 962 & 60.16 & 749 & 89.46 & 309 & 86.67 & 217 \\
                              \cmidrule{2-14}
                              & GRPO & \best{20.00} & 3017 & \underbar{20.00} & 3126 & \best{70.00} & 2008 & \best{82.60} & 1210 & \best{93.33} & 438 & \underbar{94.33} & 369 \\
                              & ARM & \underbar{16.67} & \underbar{2405} & \underbar{20.00} & \underbar{2383} & 57.50 & \underbar{1460} & 82.16 & \underbar{900} & \underbar{93.18} & \underbar{299} & 92.33 & \underbar{210} \\
                              & \textbf{Rouge} & \best{20.00} & \best{2043} & \best{23.33} & \best{2215} & \underbar{60.00} & \best{1342} & \underbar{82.20} & \best{766} & 92.33 & \best{239} & \best{95.00} & \best{161} \\
\bottomrule
\end{tabular}
\end{adjustbox}

\caption{Math benchmark results for Qwen2.5 3B/7B. \best{Bold} and \underbar{underline} indicate the best and second-best scores, respectively. For generation length (\#Tok), we only highlight the lowest and second-lowest token counts among competitive reasoning-based methods (GRPO, ARM, Rouge). SFT and O1-Pruner exhibit artificially low token counts because they often fail to perform deep reasoning; thus, their token counts are not highlighted to ensure a fair comparison of reasoning efficiency.}
\label{tab:math_result}
\end{table*}




\subsection{Training Algorithm}
\label{sec:train_algo}

Algorithm~\ref{alg:train} summarizes the overall training pipeline. 
For each question, the router allocates a shared sampling budget across 
reasoning formats, after which format-conditioned candidates are generated 
and scored using the relative efficiency reward. These rollouts are then 
used to update both the generator policy via GRPO and the router via the 
closed-form policy improvement step.
\section{Experiments}

\subsection{Experimental Setup}

\paragraph{Training data.}
Following the setup of ARM \cite{wu2025armadaptivereasoningmodel}, we utilize a mixture of \emph{three} reasoning datasets: CommonsenseQA \cite{talmor2019commonsenseqa}, GSM8K \cite{cobbe2021training}, and MATH \cite{hendrycksmath2021}. They correspond to three different difficulty levels, from simple commonsense reasoning tasks to more complex mathematical reasoning tasks.


\paragraph{Backbone.}
To evaluate across model scales, we use the public ARM checkpoints based on
Qwen2.5~3B and Qwen2.5~7B~\cite{wu2025armadaptivereasoningmodel}. Both
checkpoints can generate all reasoning formats, allowing us to isolate
improvements in format allocation and reasoning efficiency.


\paragraph{Comparison scope.}
Our target is adaptive format allocation, not generic length minimization. The
router chooses a semantic reasoning mode before generation and distributes a
fixed sampling budget across modes; length-control methods instead regulate how
long a selected trajectory continues. They provide a useful alternative
explanation for efficiency gains, but do not instantiate the same decision
problem. We therefore hold the backbone, training data, format inventory, and
rollout budget fixed and compare methods by their intervention axis.

\paragraph{Baselines.}
We select one representative for each competing explanation rather than
enumerating multiple variants of the same length penalty.
\textbf{SFT} is the no-RL control: it learns the same format tags but no adaptive
allocation policy. \textbf{GRPO}~\cite{shao2024deepseekmath} is the standard
accuracy-driven RL control and tests whether ordinary policy optimization alone
produces efficient format use. \textbf{O1-Pruner}~\cite{o1prunner} is the
representative length-control baseline; it tests whether our gains reduce simply
to rewarding shorter traces, without treating reasoning format as the action.
Finally, \textbf{ARM}~\cite{wu2025armadaptivereasoningmodel} is the closest
same-setting baseline: it uses the same multi-format action space and mitigates
format collapse with a diversity-scaled reward, but does not learn our explicit
KL-regularized, question-conditioned router. Together, these controls isolate
format routing from generic RL and token-length reduction. Additional
length-penalty variants act on the same token-level axis as O1-Pruner and do not
provide a closer test of the format-selection hypothesis.

\paragraph{Metrics.}
\begingroup
\color{black}
We evaluate correctness and efficiency using three metrics. \textbf{Pass@1} 
reports accuracy from one greedy generation per question, while \textbf{Maj@8} 
reports majority-vote accuracy over eight independently sampled generations. 
\textbf{Tokens} denotes the average completion length across generated outputs, 
serving as a proxy for inference cost.

Rouge uses the router only during 
training; at inference, each generation independently emits its own 
format-control tag as the first output token and continues in that format, 
without executing an external router.
\endgroup

\begin{table}[t!]
\centering
\small
\scriptsize
\setlength{\tabcolsep}{4pt}
\renewcommand{\arraystretch}{1.12}

\begin{adjustbox}{width=0.95\columnwidth}
\begin{tabular}{l@{\hspace{6pt}}l@{\hspace{6pt}}*{2}{r@{\hspace{6pt}}r}}
\toprule

\multirow{2}{*}{\textbf{Metric}} &
\multirow{2}{*}{\textbf{Method}} &
\multicolumn{2}{c}{\textbf{BBH}} &
\multicolumn{2}{c}{\textbf{CSQA}} \\
\cmidrule(lr){3-4}\cmidrule(lr){5-6}
& & \textbf{Score} & \textbf{\#Tok}
  & \textbf{Score} & \textbf{\#Tok} \\
\midrule

\rowcolor{gray!15}\multicolumn{6}{c}{\best{Qwen2.5 3B}}\\
\midrule

\multirow{5}{*}{\best{Pass@1}} 
                               & SFT & 37.40 & 311 & 72.80 & 99 \\
                               & O1-Pruner & \underbar{50.38} & 348 & 76.99 & 48 \\
                               \cmidrule{2-6}
                               & GRPO & 48.49 & 799 & \best{80.02} & 146 \\
                               & ARM & 48.47 & \underbar{401} & \underbar{79.12} & \underbar{59} \\
                               & \best{Rouge} & \best{50.99} & \best{262} & \best{80.02} & \best{12} \\
\midrule

\multirow{5}{*}{\best{Maj@8}} 
                              & SFT & 43.50 & 309 & 75.50 & 97 \\
                              & O1-Pruner & 52.22 & 332 & 78.05 & 48 \\
                              \cmidrule{2-6}
                              & GRPO  & \best{54.57} & 622 & \underbar{79.44} & 136 \\
                              & ARM   & \underbar{53.08} & \underbar{435} & 77.00 & \underbar{126} \\
                              & \best{Rouge} & 52.36 & \best{304} & \best{80.10} & \best{84} \\
\midrule

\rowcolor{gray!15}\multicolumn{6}{c}{\best{Qwen2.5 7B}}\\
\midrule

\multirow{5}{*}{\best{Pass@1}} 
                               & SFT & 48.20 & 245 & 80.80 & 136 \\
                               & O1-Pruner & 8.57 & 355 & 14.99 & 15 \\
                               \cmidrule{2-6}
                               & GRPO  & \best{66.36} & 639 & 84.27 & 276 \\
                               & ARM   & 63.09 & \underbar{573} & \best{84.52} & \underbar{268} \\
                               & \best{Rouge} & \underbar{64.04} & \best{407} & \underbar{84.36} & \best{10} \\
\midrule

\multirow{5}{*}{\best{Maj@8}} 
                              & SFT & 56.00 & 274 & 83.90 & 141 \\
                              & O1-Pruner & 58.44 & 305 & 28.83 & 93 \\
                              \cmidrule{2-6}
                              & GRPO  & \best{69.30} & 658 & \best{85.75} & 246 \\
                              & ARM   & \underbar{67.61} & \underbar{417} & 85.10 & \underbar{137} \\
                              & \best{Rouge} & 65.06 & \best{301} & \underbar{85.18} & \best{49} \\
\bottomrule
\end{tabular}
\end{adjustbox}

\caption{Logic and common-sense benchmark results on BBH and CSQA. \best{Bold} and \underbar{underline} indicate the best and second-best scores. We only highlight the lowest and second-lowest token counts among reasoning-based methods.}
\label{tab:logic_result}
\end{table}
\subsection{Mathematical Reasoning}
We evaluate on six mathematical reasoning benchmarks: AIME 2025 and AIME 2024 \cite{aime25,aime24}, AMC 2023 \cite{amc2023a}, MATH \cite{hendrycksmath2021}, GSM8K \cite{cobbe2021training}, and SVAMP \cite{patel2021nlp}, spanning grade-school arithmetic to Olympiad-style competition problems.

Table \ref{tab:math_result} summarizes the accuracy--efficiency comparison across both model scales. Lower generation length alone does not imply better reasoning: baseline methods such as SFT and O1-Pruner often generate fewer tokens because they underthink, leading to large accuracy drops on difficult tasks such as AIME. In contrast, compared with strong reasoning-based baselines such as GRPO and ARM, Rouge achieves a better accuracy--efficiency trade-off. Across both Qwen2.5-3B and Qwen2.5-7B, Rouge consistently produces the shortest solutions under both Pass@1 and Maj@8 while maintaining, and sometimes exceeding, baseline accuracy.

Rouge's efficiency gains are largest on easier datasets such as GSM8K and SVAMP, where long explanations are unnecessary, while remaining highly competitive on harder competition-style datasets such as AIME and AMC. Although there are a few isolated regressions, the overall results show that Rouge improves token efficiency without materially sacrificing correctness.


\subsection{General Reasoning}
We assess our approach on two general reasoning benchmarks, BIG-Bench Hard (BBH) \cite{suzgun2022challenging} and CommonsenseQA (CSQA) \cite{talmor2019commonsenseqa}, which cover diverse logic and commonsense reasoning skills. Table \ref{tab:logic_result} reports accuracy and generation length.

As in the mathematical domain, shorter outputs are only meaningful when they do not reflect underthinking. While SFT and O1-Pruner generate fewer tokens, they suffer large accuracy drops on complex logic tasks. Against GRPO and ARM, Rouge is consistently the most token-efficient across both model sizes while preserving strong general-reasoning performance. The gains are especially clear on CSQA, where Rouge matches the best baselines with far fewer tokens. On BBH, Rouge remains substantially more efficient
while staying competitive with the best baselines
on both Pass@1 and Maj@8.

\begin{figure*}[t]
  \centering
  \begin{subfigure}[t]{0.48\textwidth}
    \centering
    \includegraphics[width=\linewidth]{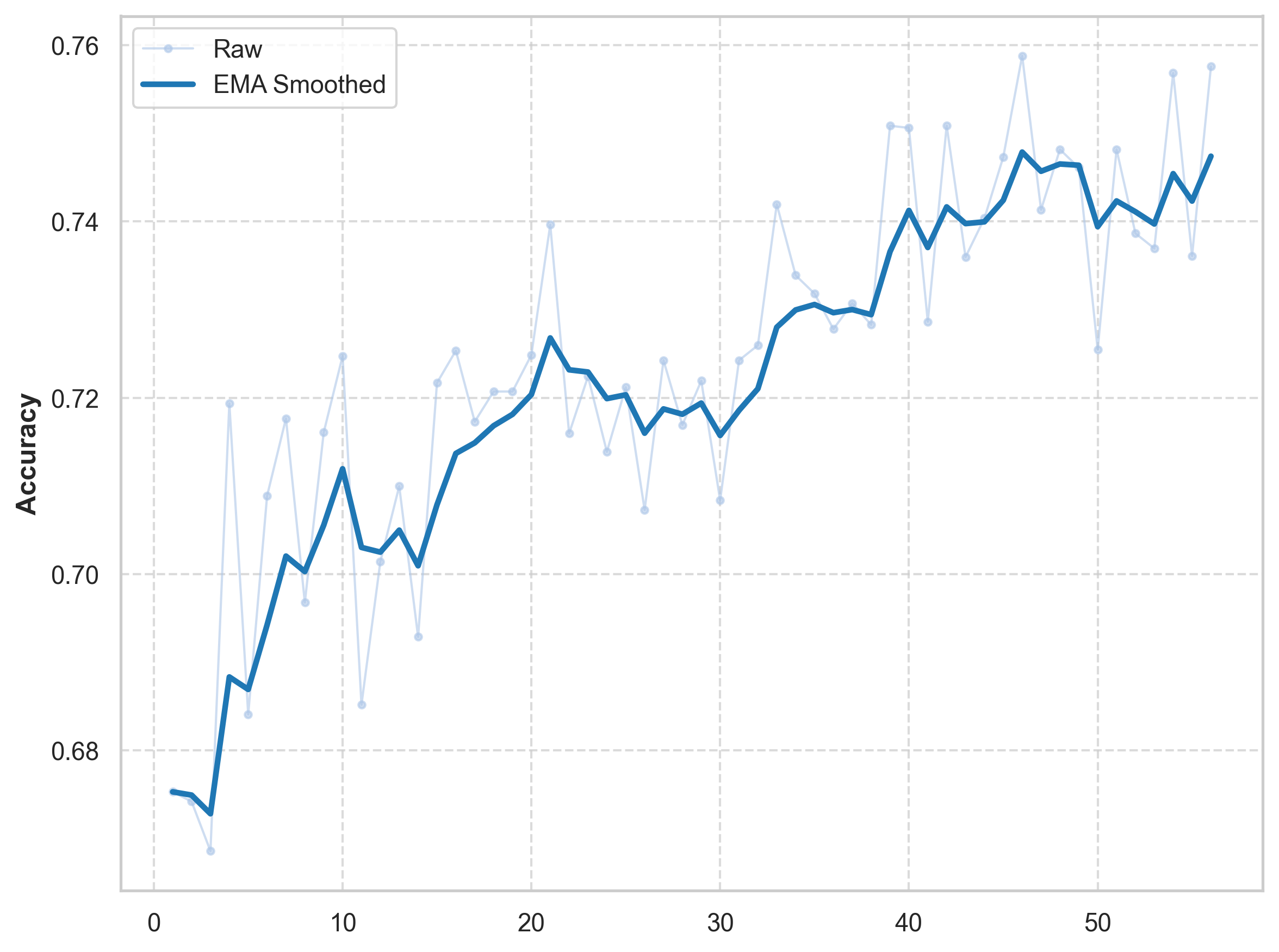}
    \caption{Training accuracy.}
    \label{fig:train_acc}
  \end{subfigure}
  \hfill
  \begin{subfigure}[t]{0.48\textwidth}
    \centering
    \includegraphics[width=\linewidth]{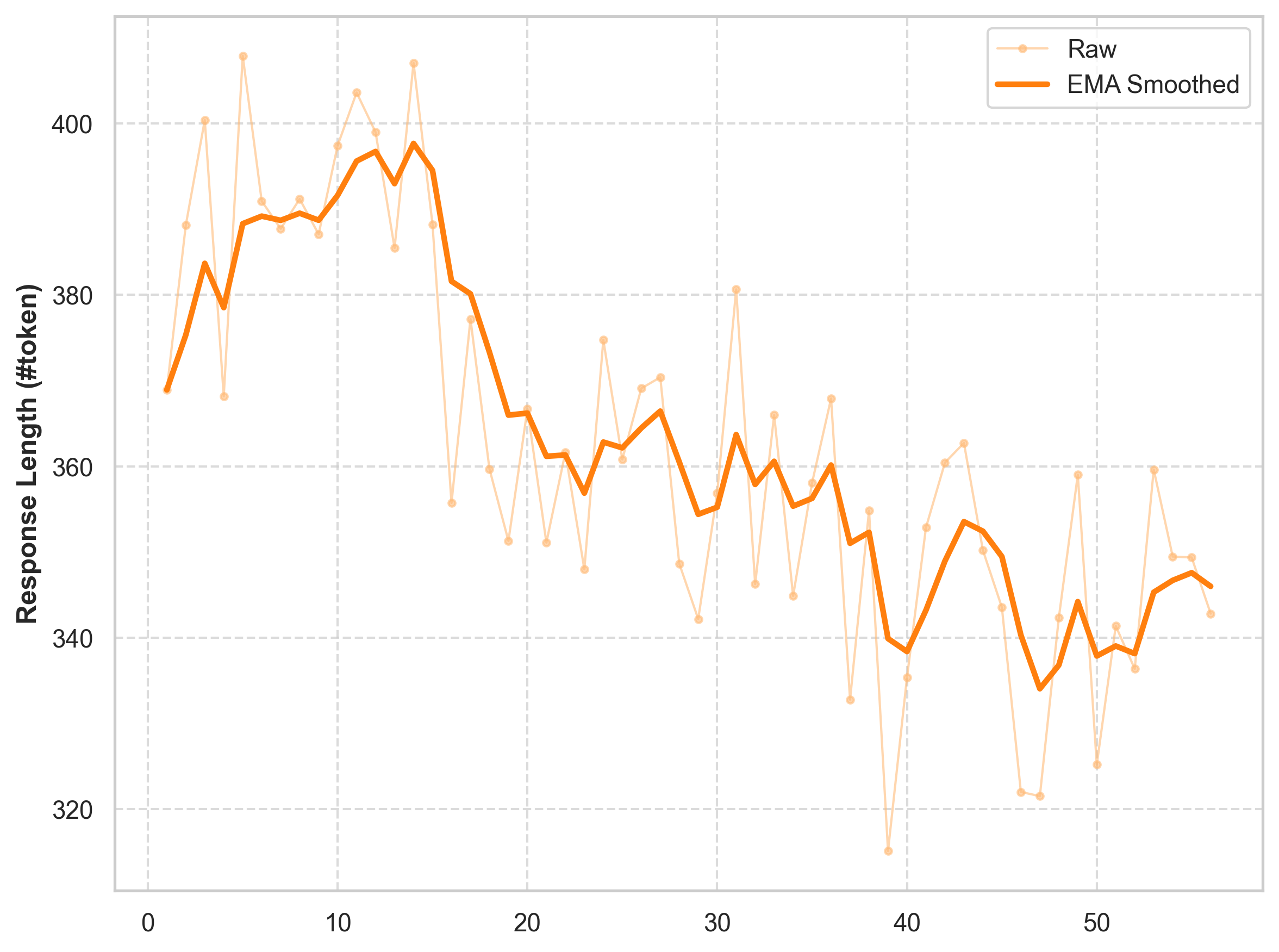}
    \caption{Training length.}
    \label{fig:train_len}
  \end{subfigure}
  \caption{Training dynamics of Rouge-7B on the training set. Accuracy rises while average response length decreases over training.}
  \label{fig:training_metrics}
\end{figure*}

\section{Analysis}

\subsection{Underthinking}
To quantify reasoning inefficiency specifically in \emph{failure cases}, we use the Underthinking metric \cite{wang2025underthinking}. Intuitively, if a model produces a correct intermediate reasoning thought early but still ends with an incorrect final answer, then the remaining tokens reflect wasted computation due to abandoning a promising path; the Underthinking score measures this wasted fraction (lower is better).

\begin{figure}[t]
  \centering
  \includegraphics[width=\columnwidth]{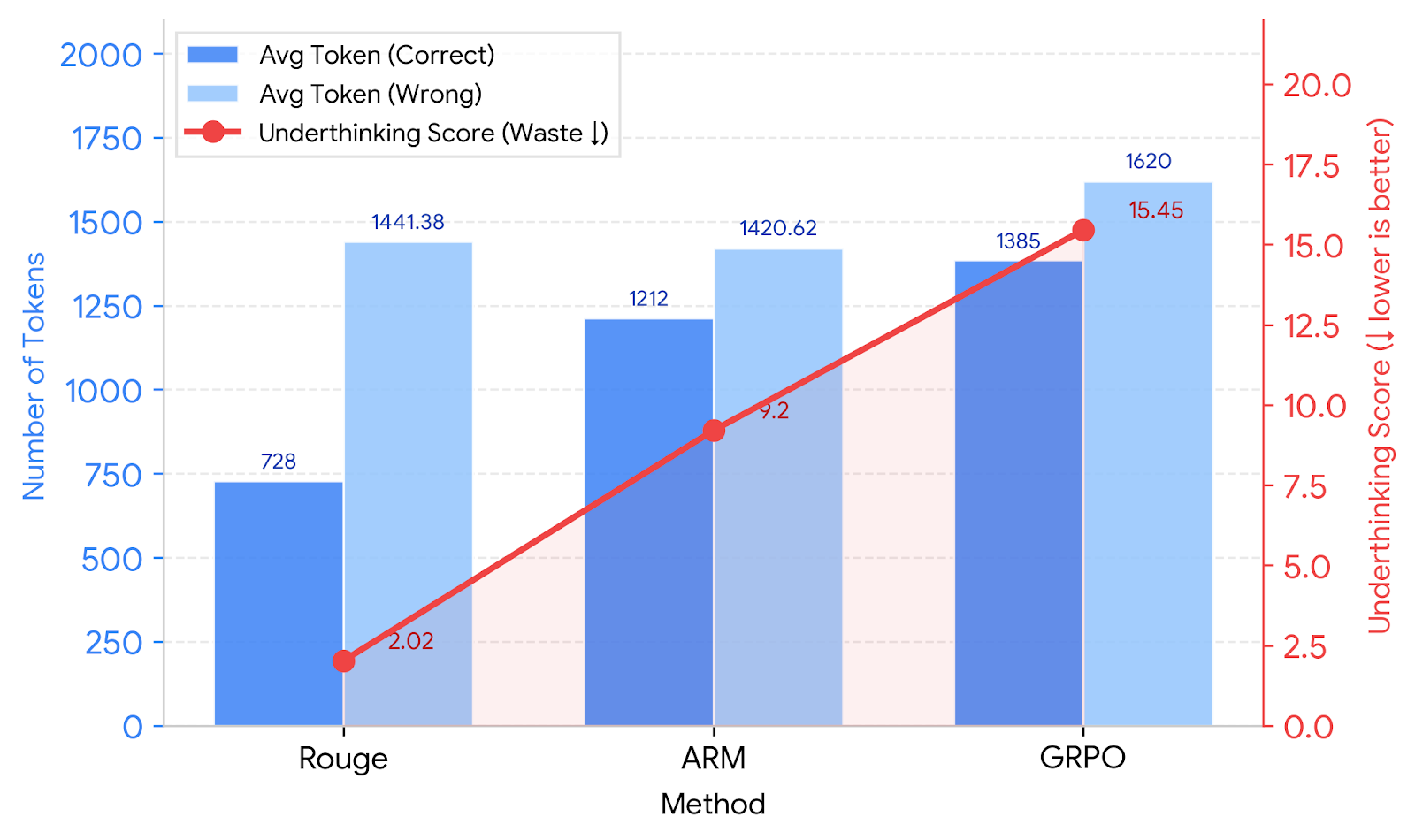}
  \caption{Generation length and underthinking waste. Rouge's significantly lower underthinking score indicates minimal wasted computation after a viable reasoning path is found compared to ARM and GRPO.}
  \label{fig:my_image}
\end{figure}

Figure~\ref{fig:my_image} shows that Rouge is both \emph{more concise when correct} and \emph{more focused when wrong}. In particular, Rouge’s underthinking waste is dramatically smaller compared to the substantially higher waste seen with ARM/GRPO. Importantly, this improvement cannot be explained purely by “stopping early”: incorrect responses are not drastically shorter, yet Rouge exhibits far less unproductive continuation after a promising reasoning path appears. By contrast, GRPO shows the strongest underthinking behavior, consistent with spending more tokens on ineffective strategy switching rather than deepening a viable line of thought.
 
\begin{figure}[t]
    \centering
    \includegraphics[width=\columnwidth]{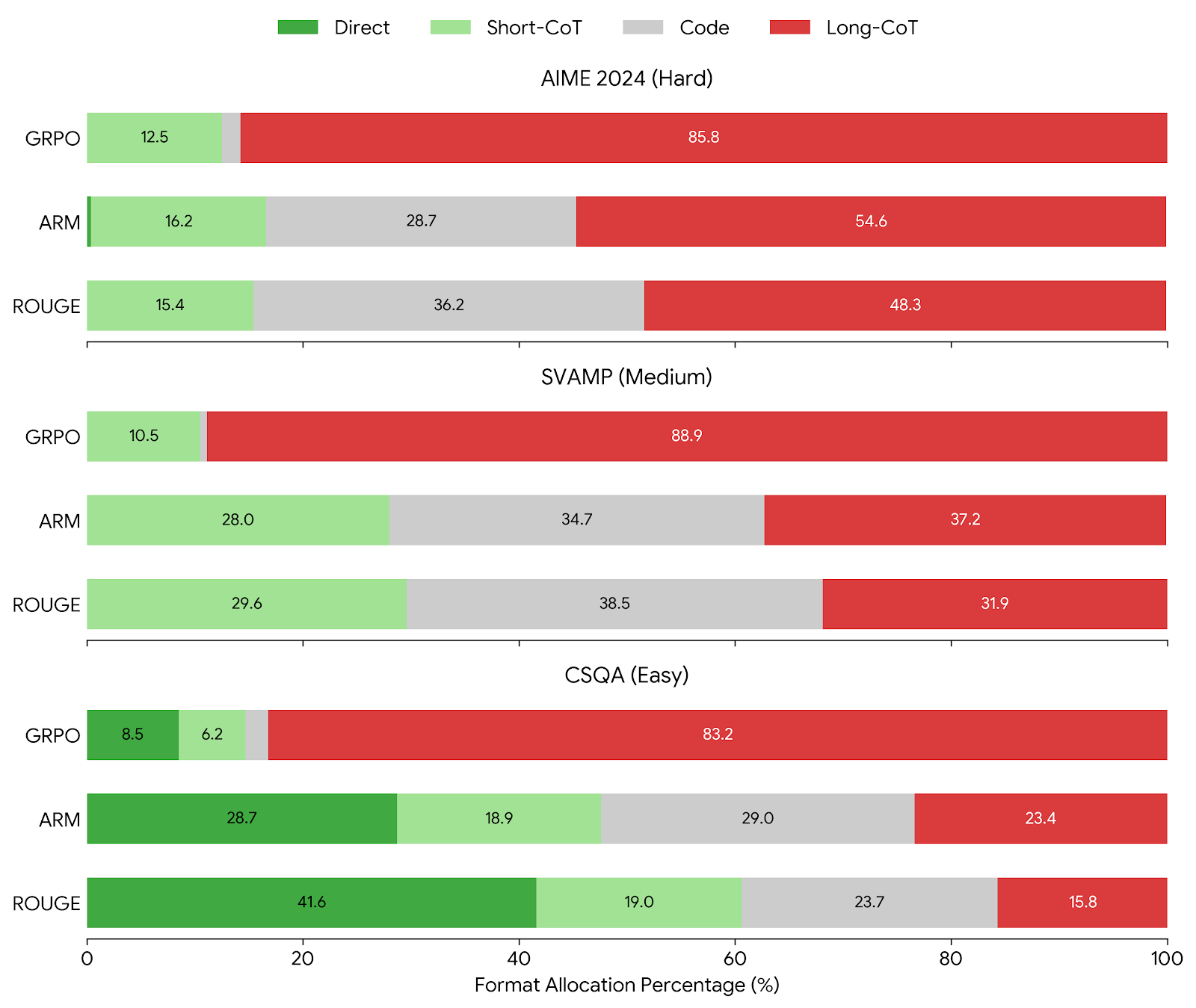}
    \caption{\textcolor{black}{Evaluation-time distribution of format-control tags emitted by each method's generated outputs; no external router is executed. Rouge uses cheaper formats more often on easy tasks, while GRPO tends to overuse \texttt{Long-CoT}.}}
    \label{fig:format_dist}
\end{figure}

\begin{figure}
    \centering
    \includegraphics[width=1\linewidth]{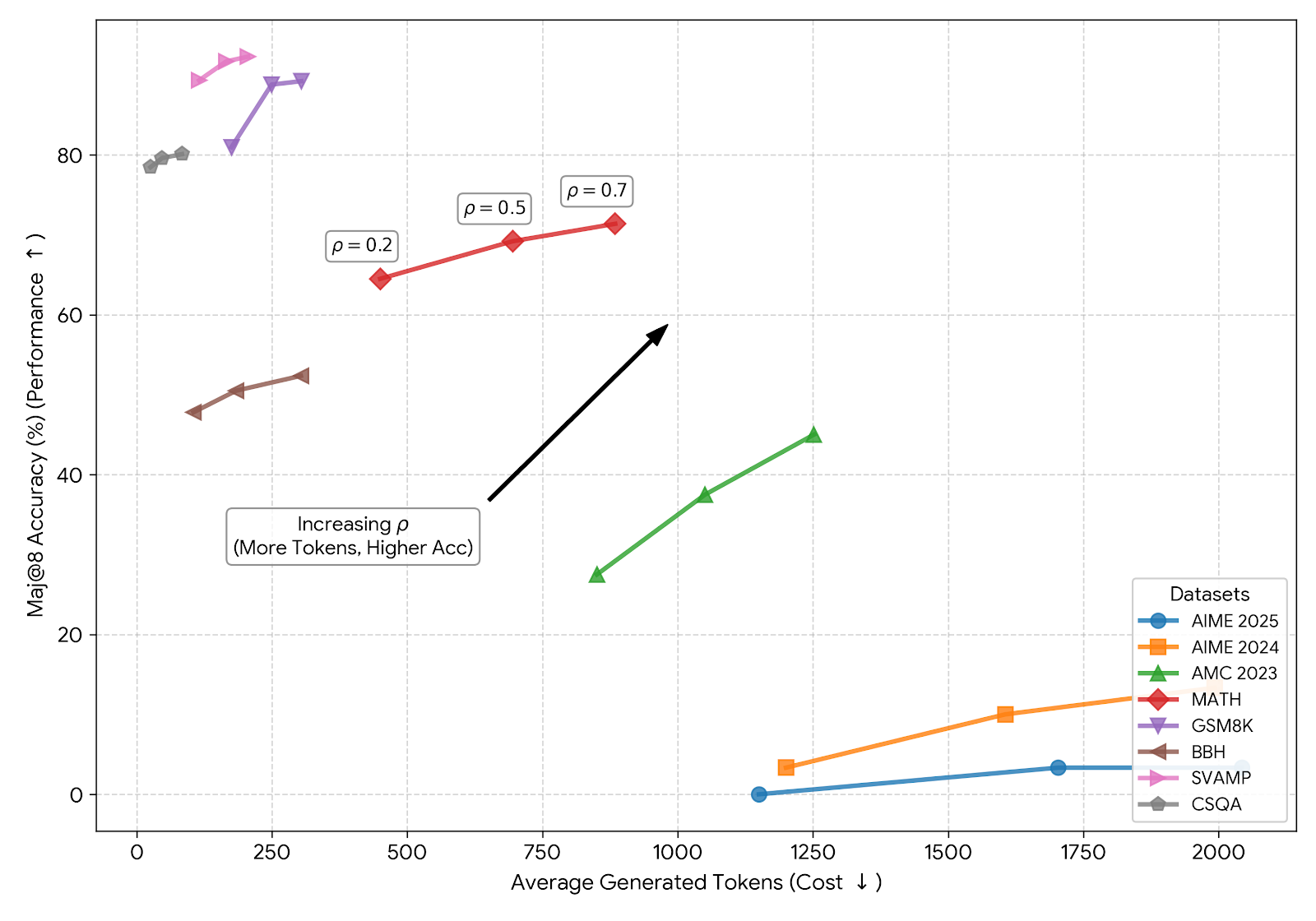}
    \caption{Sensitivity analysis of $\rho$ across benchmarks. Varying the reward margin parameter changes the accuracy--efficiency trade-off under different task difficulties.}    
    \label{fig:sen_ana}
\end{figure}

\subsection{Routing Behavior}



Figure~\ref{fig:format_dist} visualizes how each method allocates a fixed generation budget across formats (\texttt{Direct}, \texttt{Short-CoT}, \texttt{Code}, \texttt{Long-CoT}) on benchmarks of increasing difficulty. GRPO largely collapses to the most compute-heavy option, \texttt{Long-CoT}, across nearly all settings, including relatively easy questions. This suggests a miscalibrated compute policy in which expensive reasoning is overused even when simpler strategies would suffice, leading to unnecessary token consumption and lower overall efficiency.

By contrast, Rouge exhibits a more graded allocation pattern that better tracks task difficulty. On easier problems, it places more mass on cheaper formats such as \texttt{Direct} and \texttt{Short-CoT}, while on harder benchmarks it increasingly shifts toward more structured and compute-intensive strategies, including \texttt{Code} and \texttt{Long-CoT}. Compared with ARM, Rouge more consistently avoids unnecessary long reasoning on easy inputs while still allocating heavier reasoning when needed, supporting our goal of learning a principled mapping from task difficulty to the minimally sufficient reasoning format.

\subsection{Ablations}


Figure~\ref{fig:sen_ana} analyzes sensitivity to the reward margin $\rho$.
Larger values more strongly favor concise correct solutions, whereas smaller
values weaken the separation between correct and incorrect responses.
Performance changes smoothly across settings, indicating a stable
accuracy--efficiency trade-off. The preferred balance varies with task
difficulty, but no tested setting causes performance collapse.

\subsection{Training Stability}
Figure~\ref{fig:training_metrics} shows steadily improving training accuracy
while average response length decreases from roughly 390 to 345 tokens. Rouge
therefore improves performance without relying on longer generations.

\section{Conclusion}
We introduced \textbf{Route-and-Generate (Rouge)}, a training framework that decouples reasoning strategy selection from answer generation, allowing explicit control of reasoning cost in large language models. With a KL-regularized router and an efficiency-aware reward, Rouge learns to allocate computation more effectively across reasoning formats, favoring cheaper strategies on easy inputs and heavier reasoning on harder ones. Across diverse reasoning benchmarks, this yields a better accuracy--efficiency trade-off, producing shorter generations while maintaining strong performance.

\bibliography{ref}

\end{document}